\documentclass[conference]{IEEEtran}
\usepackage[latin9]{inputenc}
\usepackage{float}
\usepackage{amsmath}
\usepackage{amssymb}
\usepackage{graphicx}

\makeatletter
\newcommand{\MYfooter}{\smash{
\hfil\parbox[t][\height][t]{\textwidth}{}\hfil\hbox{}}}

\def\ps@IEEEtitlepagestyle{%
\def\@oddhead{\mbox{}2016 ICSEE International Conference on the Science of Electrical Engineering \rightmark \hfil }
\def\@oddfoot{\MYfooter{978-1-5090-2152-9/16/\$31.00\:\copyright2016\:IEEE}}%
\def\@evenfoot{\MYfooter}}
\def\ps@headings{%
\def\@oddhead{\mbox{}2016 ICSEE International Conference on the Science of Electrical Engineering \rightmark \hfil }} 
\makeatother 

\makeatother

\begin{document}

\title{Learning an Attention Model in an Artificial Visual System}

\author{\IEEEauthorblockN{Alon Hazan, Yuval Harel and Ron Meir} \IEEEauthorblockA{Department of Electrical Engineering\\
Technion \textendash{} Israel Institute of Technology \\
Technion City, Haifa, Israel} }
\maketitle
\begin{abstract}
The Human visual perception of the world is of a large fixed image
that is highly detailed and sharp. However, receptor density in the
retina is not uniform: a small central region called the fovea is
very dense and exhibits high resolution, whereas a peripheral region
around it has much lower spatial resolution. Thus, contrary to our
perception, we are only able to observe a very small region around
the line of sight with high resolution. The perception of a complete
and stable view is aided by an attention mechanism that directs the
eyes to the numerous points of interest within the scene. The eyes
move between these targets in quick, unconscious movements, known
as ``saccades''. Once a target is centered at the fovea, the eyes
fixate for a fraction of a second while the visual system extracts
the necessary information. An artificial visual system was built based
on a fully recurrent neural network set within a reinforcement learning
protocol, and learned to attend to regions of interest while solving
a classification task. The model is consistent with several experimentally
observed phenomena, and suggests novel predictions. 
\end{abstract}

\section{Introduction}

% no \IEEEPARstart
Neuroscientists and cognitive scientists have many tools at their
disposal to study the brain and neural networks in general, including
Electroencephalography (EEG), Single-Photon Emission Computed Tomography
(SPECT), functional Magnetic Resonance Imaging (fMRI) and Microelectrode
Arrays (MEA), to name a few. However, the amount of information and
level of control afforded by these tools do not remotely resemble
what is available to an engineer working on an artificial neural network.
The engineer can manipulate any neuron at any time, force certain
excitations, intervene in ongoing processes, and collect as much data
about the network as needed, at any level of detail. This wealth of
information has enabled reverse engineering research on artificial
neural networks, leading to insights into the inner workings of trained
artificial neural networks. This suggests an indirect approach to
studying the brain: training a biologically plausible neural network
model to exhibit complex behavior observed in real brains, and reverse
engineering the result. In line with this approach, we designed an
artificial visual system based on a fully recurrent unlayered neural
network that learns to perform saccadic eye movements. Saccadic eye
movements are quick, unconscious, task-dependent \cite{Yarbus1967a}
motions following the demand of attention \cite{Buswell1935a}, that
direct the eye to new targets that require the high resolution of
the fovea. These targets are usually detected within the peripheral
visual system \cite{Pelz2000a}. Once a target is centered at the
fovea, the eye fixates for a fraction of a second while the visual
system extracts the necessary information. Most eye movements are
proactive rather than reactive, predict actions in advance and do
not merely respond to visual stimuli \cite{Land1997a}.

There is good evidence that much of the active vision in humans results
from Reinforcement Learning (RL) \cite{Schultz2000a}, as part of
and organism's attempt to maximize its performance while interacting
with the environment \cite{Hayhoe2005b}. Accordingly, we train the
artificial visual system within the RL paradigm. The network was not
explicitly engineered to perform a certain task, and does not contain
an explicit memory component \textemdash{} rather it has memory only
by virtue of its recurrent topology. Learning takes place in a model-free
setting using policy gradient techniques. 

We find that the network displays attributes of human learning such
as: (a) decision making and gradual confidence increase along with
accumulated evidence, (b) skill transfer, namely the ability to use
a pre-learned skill in a certain task in order to improve learning
on a related but more difficult task, (c) selectively attending information
relevant for the task at hand, while ignoring irrelevant objects in
the field of view.

\section{\label{sec:The-artificial-visual}The artificial visual system}

We designed an Artificial Visual System (AVS) with the task of learning
an attention model to control saccadic eye movements, and subsequent
classification of digits. We refer to this task as the \emph{attention-classification
task}. The AVS is similar in many ways to that presented in \cite{Mnih2014}.
It is a simplified model of the human visual system, consisting of
a small region in the center with high resolution, analogous to the
human fovea, and two larger concentric regions which are sub-sampled
to lower resolution and are analogous to the peripheral visual system
in humans. The AVS was trained and tested on the classification of
handwritten digits from the MNIST data set \cite{LBBH98} . Only a
small part of the image is visible to the AVS at any one time. Specifically,
full resolution is only available at the fovea, which is 9-by-9 pixels,
as in \cite{Mnih2014}, or 5-by-5 pixels (about 69\% smaller). The
first peripheral region is double the size of the fovea, but sub-sampled
with period 2 to match the size of the fovea in pixels. Similarly,
the second peripheral region is quadruple the size of the fovea but
sub-sampled with period 4. For comparison, a typical digit in the
MNIST database occupies about 20-by-20 pixels of the image. The location
of the observation within the image is not available to the AVS (unlike
\cite{Mnih2014}), and movements of the observation location are not
constrained to image boundaries. Instead, locations outside the image
boundaries are observed as black pixels.

\begin{figure}[h]
\centering \includegraphics[scale=0.35]{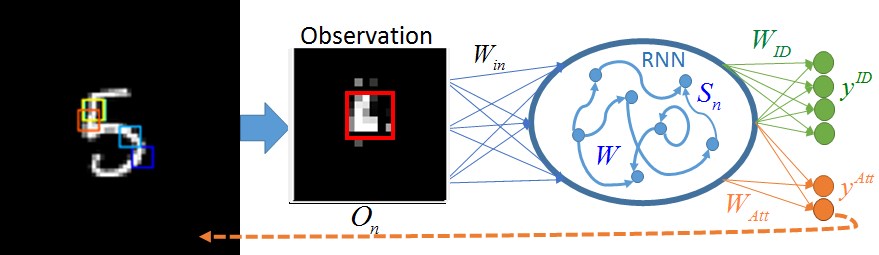}

\caption{Artificial Visual System design \label{fig:Artificial-Visual-System}}
\end{figure}

The AVS consists of inputs (observations) projected upon the network
via input weights $W_{\mathrm{in}}$, a neural network consisting
of $N$ neurons connected by the recurrent weights $W$, and two outputs:
a classifier $y^{\mathrm{ID}}$, responsible for identifying the digit
after the AVS has explored the image, and the attention model output
$y^{\mathrm{Att}}$, responsible for directing the eye to new locations
based on the information represented in the network state (see Figure
\ref{fig:Artificial-Visual-System}). The output $y_{\mathrm{ID}}$
consists of one neuron for each possible digit. At the end of the
trial, the identity of the highest valued neuron is interpreted as
the network's classification. The progression of a single trial follows
these principal stages: 
\begin{enumerate}
\item A random digit is selected from the MNIST training database. 
\item A location across the image is randomly selected.
\item The observation (called `glimpse' \cite{Mnih2014}) from the current
location is projected upon the network through $W_{\mathrm{in}}$,
along with any pre-existing information within the network state through
the recurrent weights $W$. 
\item The attention model output $y^{\mathrm{Att}}$ is fed back as a saccade
of the eye, i.e. as the size of movement from the current location
in the horizontal and vertical axes. 
\item If a predefined number of glimpses has passed (or by network decision),
compare the classifier output $y^{\mathrm{ID}}$ to the true label,
otherwise return to stage 3. 
\item Reward the AVS if the classification was correct, and continue to
the next trial. 
\end{enumerate}
The AVS is implemented by a fully recurrent neural network. Its network
topology is similar the Echo State Network (ESN) \cite{Jaeger2001}
in that the recurrent neural connections are drawn randomly and are
not constrained to a particular topology such as in layered feedforward
networks, or long short-term memory networks.

The network state evolves according to

\begin{align}
s_{n+1} & =\left(1-\alpha\right)s_{n}+\alpha\left(W\tanh\left(s_{n}\right)+W_{\mathrm{in}}o_{n+1}+\zeta_{n}\right),\nonumber \\
y_{n} & =W_{\mathrm{out}}\tanh\left(s_{n}\right)+\xi_{n},\label{eq:avs}
\end{align}
where 
\begin{itemize}
\item $s_{n}\in\mathbb{R}^{N}$ is the state of network at time step $n$,
each element representing the state of a single neuron, 
\item $\alpha\in(0,1]$ is the leak rate,
\item $W\in\mathbb{R}^{N\times N}$ is the internal connections weight matrix,
\item $W_{\mathrm{in}}\in\mathbb{R}^{N_{\mathrm{in}}\times N}$ is the input
weight matrix,
\item $o\in\mathbb{R}^{N_{\mathrm{in}}}$ is the observation (network input),
\item $\zeta_{n}\in\mathbb{R}^{N}$ and $\xi_{n}=\left(\xi_{n}^{\mathrm{ID}};\xi_{n}^{\mathrm{Att}}\right)\in\mathbb{R}^{M}$
are independent discrete-time Gaussian white noise processes with
independent components, each having variance $\sigma_{\zeta}^{2},\sigma_{\xi}^{2}$
respectively,
\item $y_{n}=\left(y_{n}^{\mathrm{ID}};y_{n}^{\mathrm{Att}}\right)\in\mathbb{R}^{M}$
is the state of the $M$ output neurons, 
\item $W_{\mathrm{out}}\in\mathbb{R}^{M\times N}$ is the output weight
matrix (consisting of blocks $W_{\mathrm{ID}},W_{\mathrm{Att}}$ for
the corresponding output components).
\end{itemize}
The gradient of the expected reward $J$ is estimated as in \cite{Willia1992},
\[
\hat{\nabla}J=\left\langle \nabla\log p\left(\tau\right)\left(r\left(\tau\right)-b\right)\right\rangle ,
\]
where $\tau=(s_{0},w_{1},(o_{n},s_{n},y_{n}^{\mathrm{Att}},y_{n}^{\mathrm{ID}})_{n=1}^{N_{g}})$
is a random trajectory of $N_{g}$ glimpses, $p\left(\tau\right)$
is the probability of trajectory $\tau$, $r\left(\tau\right)$ the
observed (usually binary) reward, $b$ a fixed baseline computed as
in \cite{Willia1992}, $w_{1}$ is the random location of the first
glimpse, and $\left\langle \cdot\right\rangle $ indicates averaging
over trajectories. Viewed as a partially observable Markov decision
process (POMDP), we can write the distributions describing the agent:
\begin{align}
p\left(s_{n+1}|s_{n},o_{n+1}\right) & =\alpha^{-1}p_{\zeta}\Big(\alpha^{-1}\left(s_{n+1}-\left(1-\alpha\right)s_{n}\right)\nonumber \\
 & \qquad\qquad-W\tanh\left(s_{n}\right)-W_{\mathrm{in}}o_{n+1}\Big)\nonumber \\
p\left(y_{n}|s_{n}\right) & =p_{\xi}\left(y_{n}-W_{\mathrm{out}}\tanh\left(s_{n}\right)\right),\label{eq:p-out}
\end{align}
where $p_{\zeta},p_{\xi}$ are the probability density functions of
$\zeta_{t},\xi_{t}$ respectively. The POMDP dynamics are deterministic:
the glimpse position $w_{n}$ evolves as $w_{n+1}=w_{n}+y_{n}^{\mathrm{Att}}$.

For the AVS \eqref{eq:avs} the probability density of a trajectory
$\tau$ is
\begin{align*}
p\left(\tau\right)=p\left(s_{0}\right)p\left(w_{1}\right)\prod\limits _{n=1}^{N_{g}} & p\left(s_{n+1}|s_{n},o_{n+1}\right)p\left(y_{n+1}|s_{n+1}\right),
\end{align*}
Here only the output probabilities $p\left(y_{n+1}|s_{n+1}\right)$
depend on $W_{\mathrm{Out}}$, and, using \eqref{eq:p-out} and the
Gaussian distribution of the noise, we find that the log likelihood
gradient with respect to $W_{\mathrm{Out}}$ takes the form
\begin{align*}
\nabla_{W_{\mathrm{Out}}}\log p\left(\tau\right) & =\sum\limits _{n=1}^{N_{g}}\sigma_{\xi}^{-1}\xi_{n}\tanh\left(s_{n}\right)^{T},
\end{align*}
where $N_{g}$ is the number of glimpses. Stochastic gradient ascent
is performed only for the output weights $W_{\mathrm{Out}}$. Recurrent
weights are randomly selected, with spectral radius 1, and remain
fixed throughout training. The log likelihood with respect to the
internal weight matrix $W$ takes a similar form. However, the recurrent
connections were not learned in our simulations.

\section{Results}

\subsection{\label{sub:Exploitation-of-memory}Use of memory}

Since information accumulated by the neural network over time is mixed
into the state of the network, it is not obvious that the potential
to extract useful historic information can be exploited within the
attention model solution. Training uses gradient ascent to the local
maxima of the estimated expected reward and therefore may converge
to sub-optimal maxima that do not make use of the full potential of
the system. In order to test the use of memory by the trained network,
two similar AVS were trained on the attention-classification task.
In the first AVS, recurrent weights were random, whereas the second
AVS was set to `forget' historic information, by setting the recurrent
weights matrix to zero.

\begin{figure}[h]
\centering \includegraphics[scale=0.4]{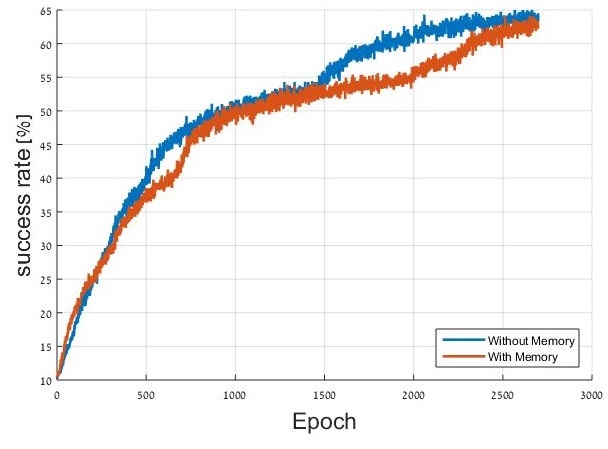}

\caption{Exploitation of memory with 81px fovea\label{fig:Exploitation-of-memory-R5}}
\end{figure}

Use of memory was found to depend on the size of the fovea. Fig. \ref{fig:Exploitation-of-memory-R5}
shows the performance of the system across training epochs, for the
case of a large (9$\times$9 pixels) fovea. Initially, the AVS with
memory has the advantage as the attention model is still poor at this
stage, leading to relatively uninformative glimpses, so the use of
information from several glimpses results in better classification.
However, as the attention mechanism improves the last glimpse becomes
highly informative, so the memoryless network, where information from
the last glimpse is not corrupted by memory of previous glimpses,
has the advantage. In fact, we found that information from a well-placed
glimpse suffices to classify the digit with over 90\% success rate
in this case, driving the network to a solution of finding a single
good glimpse location across the digit, and classification based on
that glimpse, without regard to the rest of the trajectory.

\begin{figure}[h]
\centering \includegraphics[scale=0.4]{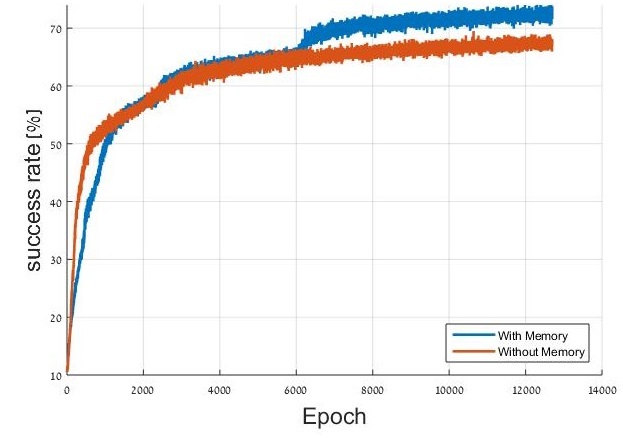}
\caption{Exploitation of memory with 25px fovea\label{fig:Exploitation-of-memory-R3}}
\end{figure}

The situation is different with a smaller fovea (5$\times$5 pixels),
where classification from a single glimpse becomes harder. As seen
in Figure \ref{fig:Exploitation-of-memory-R3}, the AVS with memory
outperforms the one without memory in the small fovea case. 

\subsection{\label{sub:Gathering-Information}Gathering Information }

The human visual system acts to maximize the information relevant
to the task \cite{Henderson2003a}. In order to assess whether our
AVS behaves similarly, we have to characterize the relevant information
in the context of our task. Since the network classification $y^{\mathrm{ID}}$
depends linearly on the network state in the last time step, we quantify
the task-relevant information as the best linear separation of the
network state, between each class and the other classes. Accordingly,
we use Linear Discriminant Analysis (LDA) \cite{Fukunaga1990}, which
acts to find the projection that minimizes the distance between samples
of the same cluster $S_{w}$ while at the same time maximizes the
distance between clusters $S_{b}$. The distance within each class
is measured by the variance of samples belonging to that class, and
$S_{w}$ is taken to be the mean of these distances across all classes.
The Distance between classes $S_{b}$ is defined as the variance of
the set of class centers. 

\begin{figure}[h]
\centering \includegraphics[scale=0.47]{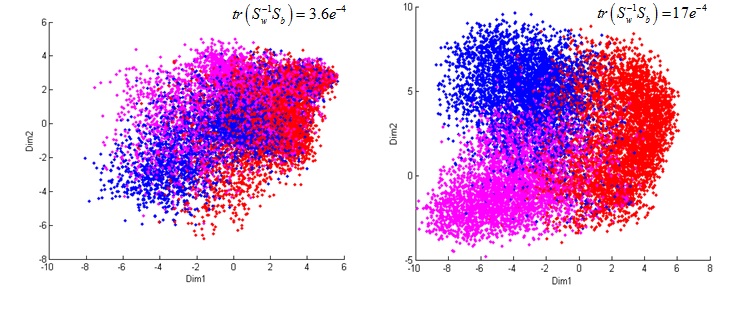}
\caption{\label{fig:Gathering-Information}Results of Linear Discriminant Analysis
of the AVS state at the last time step. Each dot corresponds to single
trial and represents the projection of the network state on the first
two eigenvectors of $S_{w}^{-1}S_{b}$. Dots are colored according
to the digit presented to the network. Left: random last glimpse.
Right: full attention model.}
\end{figure}

We trained an AVS on the attention-classification task with 5 glimpses
per digit. After the AVS was trained, it was tested in two cases.
In the first case, the system was run as usual and the network state
vector was recorded after the last glimpse of each digit in the test
set. In the second case, the location of the last glimpse was chosen
randomly rather than following the learned attention model. The results
are illustrated in Figure \ref{fig:Gathering-Information}, where
the state of the network is projected on the first two eigenvectors
of $S_{w}^{-1}S_{b}$. Separation is significantly better with the
full attention model compared to the one with the random last glimpse.
We conclude that, at the very least, the attention model acts to maximize
task-relevant information in the last glimpse better than a random
walk.

\subsection{\label{sub:Transfer-learning-AVS}Transfer learning }

\begin{figure}[h]
\centering \includegraphics[scale=0.4]{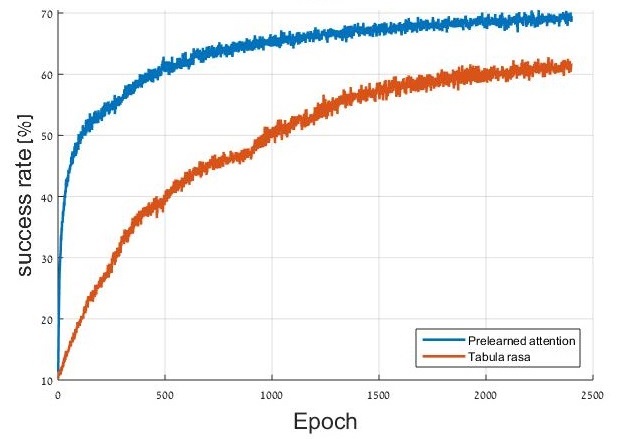}
\caption{Transferable Skill\label{fig:Transferable-Skill}}
\end{figure}

Biological learning often displays the ability to use a skill learned
on a simple task in order to improve learning of a harder yet related
task, e.g., proficiency at tennis is beneficial when learning racquetball
and even seemingly unrelated tasks such as skiing for example \cite{Seidler2010}.
To test whether transfer learning is possible in the AVS, we trained
it to learn the attention model and classification of 3 digits (out
of 10 in the MNIST database). The resulting solution served as an
initial condition for learning the full task of classifying all 10
digits. As seen in Fig. \ref{fig:Transferable-Skill}, not only did
the AVS with pre-learned attention learn much faster, but it also
achieved a better result at the end of training.

\subsection{\label{sub:Ignoring-distractions}Ignoring distractions}

The eyes are not directed to the most visually salient points in the
field of view, but rather to the ones that are most relevant for the
task at hand \cite{Hayhoe2005b}. Accordingly, we introduced an highly
salient object into the training images. The object is a square, approximately
the size of a digit but with maximum brightness, whereas the digits
are handwritten and displayed in grayscale. The object is inserted
at a fixed position relative to the digit, always on the right hand
side of the digit. The trained network successfully avoids unnecessary
fixations on the salient object. In cases where the first glimpse
falls upon an area where both the digit and the object are within
the peripheral visual region, the object seems to be completely ignored.
Perhaps more interesting is the case where only the object is visible
in the first glimpse, within the peripheral view. In such a case,
the AVS learned to exploit the fact that the digit is always located
on the left of the object and consistently performs saccades to the
left. Thus, not only was the presence of a distracting object not
harmful to performance, but it was actually beneficial.

\begin{figure}[h]
\centering \includegraphics[scale=0.7]{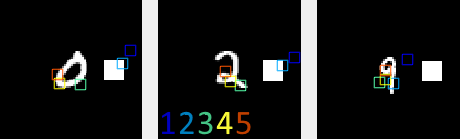}
\caption{Fixed Distracting Object\label{fig:Fixed-Distracting-Object}}
\end{figure}

In Fig. \ref{fig:Fixed-Distracting-Object}, the colored squares represent
the foveal view of 5-by-5 pixels at each time step going from blue
to red. The left and middle images show cases where the first glimpse
only observes the distracting object within the peripheral view. The
left image shows a case where the first glimpse observes both the
distracting object and the digit within the peripheral view.

Next, we test the network with a distracting object in a random position
around the digit. The observed behavior was similar when the first
glimpse happened to fall on a location where both the object and digit
are within the peripheral view: the AVS ignored the distraction and
directed itself towards the digit. However, in the case where the
first glimpse falls on a location where only the distracting object
is visible in the peripheral view, the AVS failed to locate the digit. 

\begin{figure}[h]
\centering \includegraphics[scale=0.35]{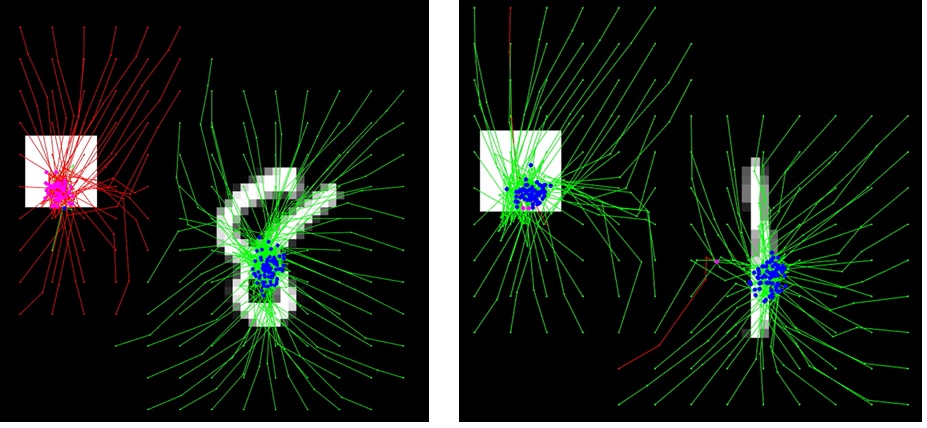}
\caption{Free Distracting Object\label{fig:Free-Distracting-Object}}
\end{figure}

An example is seen in Fig. \ref{fig:Free-Distracting-Object}. The
lines are trajectories of the AVS each starting from a different point
on a test grid and followed until the last glimpse (blue/magenta dot).
Green lines are trajectories that led to a correct classification
while red lines are trajectories that led to a false classification.
When the AVS happen to fall at a location where the digit is not seen,
it directs its gaze towards the square, which it then chooses to classify
as ``1'' thus earning 10\% expected reward which is better than
nothing.

\subsection{\label{sub:Learning-aided-by}Learning aided by demonstration (guidance)}

\begin{figure}[H]
\centering \includegraphics[scale=0.4]{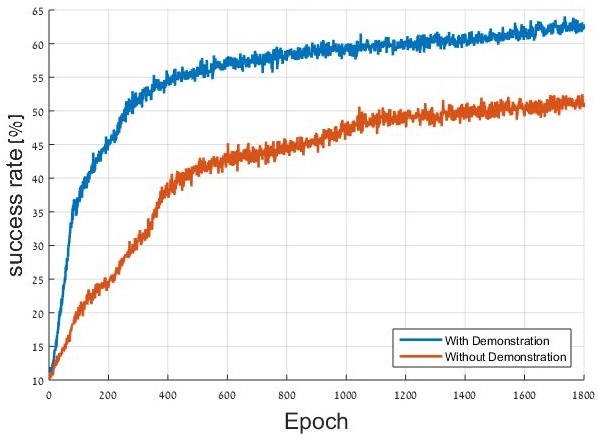}
\caption{Demonstration \label{fig:Demonstraion-Learning}}
\end{figure}

Learning by demonstration (or learning with guidance) was implemented
in the AVS. Demonstration differs from supervision in two key ways.
First, demonstration is not continuous, and is applied sparsely in
time in order to suggest new trajectories to the system. Second, demonstration
is not required to provide the best solution to the system, because
the system maintains its freedom to explore and even improve upon
it. Demonstration was achieved by providing the network with a sparse
and naive suggestion for the attention model. For example, on $10\%$
of trajectories, the system was directed to the center of the digit
on the last glimpse. Such partial direction resulted in a significant
improvement of both speed of learning and the final success rate,
as can be observed in figure \ref{fig:Demonstraion-Learning}. 

Demonstration in the AVS system was made possible by manipulating
the exploration noise. The exploration noise is a Gaussian white noise
and as such has probability greater than zero to accept any value.
Since the output of the system at any given time is a function of
that noise, we can force the output to a specific value by setting
the exploration noise in that particular time step to be ${\tilde{\zeta}_{n}}=y_{n}^{\mathrm{att}}-{W_{\mathrm{Att}}}\tanh\left(s_{n}\right)$
where $y_{n}^{\mathrm{att}}$ is now the demonstrated output of the
attention model and $\tilde{\zeta}_{n}$ is the determined exploration
noise that will bring the system to that desired output. As long as
the demonstration is kept sparse enough, it would in practice not
break the assumption that the noise is a Gaussian white noise. The
noise in the system is an essential part of the log likelihood gradient
and therefore the system would not only arrive to the desired output
at that particular time step, but also learn from that experience.

\section{Conclusion}

We have shown that a simple artificial visual system, implemented
through a recurrent neural network using policy gradient reinforcement
learning, can be trained to perform classification of objects that
are much larger than its central region of high visual acuity. While
receiving only classification based reward, the system develops an
active vision solution, which directs attention towards relevant parts
of the image in a task-dependent way. Importantly, the internal network
memory plays an essential role in maintaining information across saccades,
so that the final classification is achieved by combing information
from the current visual input and from previous inputs represented
in the network state. Within a generic active vision system, without
any specifically crafted features, we have been able to explain several
features characteristic of biological vision: \emph{(i)} Good classification
performance using reinforcement learning based on highly limited central
vision and low resolution peripheral vision, \emph{(ii)} Gathering
task-relevant information through active search, \emph{(iii)} Transfer
learning, \emph{(iv)} Ignoring task-irrelevant distractors, \emph{(v)}
Learning through guidance. Beyond providing a model for biological
vision, our results suggest possible avenues for cost-effective image
recognition in artificial vision systems. 

The Matlab code will be made available at: https://alonhazan.wordpress.com/

\end{document}